\documentclass[10pt,twocolumn,letterpaper]{article}

\usepackage[pagenumbers]{cvpr} % To force page numbers, e.g. for an arXiv version

\definecolor{cvprblue}{rgb}{0.21,0.49,0.74}
\usepackage[pagebackref,breaklinks,colorlinks,allcolors=cvprblue]{hyperref}

\def\paperID{} % *** Enter the Paper ID here
\def\confName{CVPR}
\def\confYear{2026}

\usepackage[T1]{fontenc}
\usepackage{hyperref}
\usepackage{url}
\usepackage{graphicx}
\usepackage{booktabs}
\usepackage{multirow}
\usepackage{makecell}
\usepackage{amsmath}
\usepackage{minted}
\usepackage{soul}
\usepackage{listings}
\definecolor{dkgreen}{rgb}{0,0.6,0}
\definecolor{gray}{rgb}{0.5,0.5,0.5}
\definecolor{mauve}{rgb}{0.58,0,0.82}
\usepackage[table]{xcolor}
\usepackage{ulem}
\usepackage{float}

\title{Physically Ground Commonsense Knowledge for Articulated Object Manipulation with Analytic Concepts}

\author{
Jiude Wei$^{1}$, ~~~ Yuxuan Li$^{2}$, ~~~  Cewu Lu$^{1,3}$, ~~~ Jianhua Sun$^{1}$\footnotemark[2] \\
\small{$^1$School of Artificial Intelligence, Shanghai Jiao Tong University} ~~~ \small{$^2$School of Computer Science, Shanghai Jiao Tong University}\\ 
\small{$^3$Shanghai Innovation Institute}
}

\begin{document}
\maketitle

\renewcommand{\thefootnote}{\fnsymbol{footnote}}
\footnotetext[2]{denotes corresponding author}

\begin{abstract}
We humans rely on a wide range of commonsense knowledge to interact with an extensive number and categories of objects in the physical world. Likewise, such commonsense knowledge is also crucial for robots to successfully develop generalized object manipulation skills. While recent advancements in Multi-modal Large Language Models (MLLMs) have showcased their impressive capabilities in acquiring commonsense knowledge and conducting commonsense reasoning, effectively grounding this semantic-level knowledge produced by MLLMs to the physical world to thoroughly guide robots in generalized articulated object manipulation remains a challenge that has not been sufficiently addressed. To this end, we introduce analytic concepts, procedurally defined upon mathematical symbolism that can be directly computed and simulated by machines. By leveraging the analytic concepts as a bridge between the semantic-level knowledge inferred by MLLMs and the physical world where real robots operate, we can figure out the knowledge of object structure and functionality with physics-informed representations, and then use the physically grounded knowledge to instruct robot control policies for generalized and accurate articulated object manipulation. Extensive experiments in both real world and simulation demonstrate the superiority of our approach. 
\end{abstract}

\section{Introduction}

Studies in human cognition~\citep{carey2001infants,leslie1998indexing,ullman2000high,biederman1987recognition,hummel1992dynamic} figure out that humans possess a wide range of commonsense knowledge which guides us to properly interact with an extensive number and categories of objects in the physical world. Such knowledge spans multiple physical concepts, including the spatial structures, functional properties and physical dynamics of objects. We are able to ground the commonsense knowledge from human cognition to the physical world and effectively manipulate real-world objects based on this knowledge to accomplish specific tasks. 

Commonsense knowledge is also essential for the successful development of general robot manipulation. As recent advancements in MLLMs have shown substantial potential in utilizing commonsense knowledge and performing commonsense reasoning~\citep{achiam2023gpt,hurst2024gpt,liu2024visual,touvron2023llama}, researchers have leveraged MLLMs in articulated object manipulation to pursue more generalized manipulation capabilities~\citep{li2024manipllm,tang2023graspgpt,huang2024manipvqa,3dllm,kim2025openvla,ghosh2024octo}. 
MLLM-instructed paradigm incorporates an understanding of the semantics underlying the manipulation tasks~\citep{karamcheti2023language,schmidtlearning,nair2023r3m}. This facilitates an intuitive comprehension of the task goal, and enabling the semantic-level reasoning between actions and outcomes, which guides the robot to reasonably complete manipulation tasks, especially for open-ended tasks in the real-world environment~\citep{bender2020climbing,huang2023voxposer}.

\begin{figure*}[htbp]
    \centering
    \includegraphics[width=\linewidth]{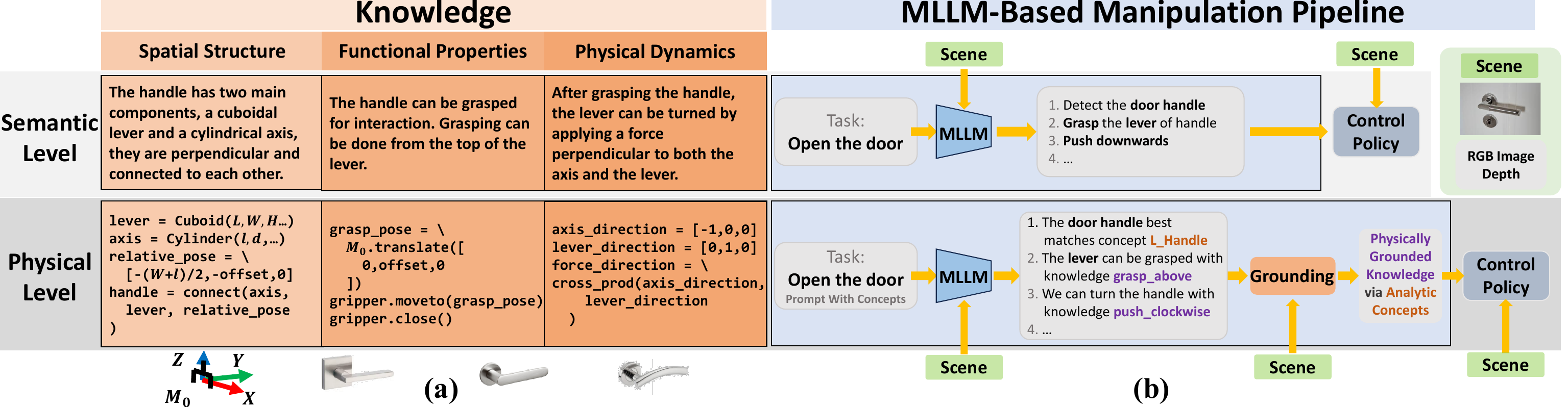}
    \vspace{-15pt}
    \caption{\textbf{(a):} An illustration of knowledge represented using both semantic and physical expressions. Compared to semantic expressions, physical ones provide precise mathematical definitions and numerical values, which robots can more effectively interpret and compute for object manipulation (detailed in \cref{fig:analytic-concepts}-Left). 
    \textbf{(b):} Illustrations of the conventional MLLM-driven object manipulation pipeline (Upper) and our approach (Lower). Using analytic concepts, we bridge the gap between semantic-level representations, which MLLM excels at, and the physical world, on which robots operate. Please refer to \cref{fig:analytic-concepts} for descriptions of \textbf{L\_Handle}, \textbf{grasp\_above} and \textbf{push\_clockwise}.}
    \label{fig:phy-knowledge-pipeline}
    \vspace{-12pt}
\end{figure*}

However, since LLMs function at the semantic level, significant challenges still remain in effectively integrating their commonsense reasoning capability into the control policy, which operates at the physical level to manage real-robot interactions. On one hand, encoding knowledge in natural language as feature inputs for the control policy may hinder the control policy from fully recognizing the physical concepts underlying the knowledge~\citep{majumdar2023we, cohen2024survey}. On the other hand, LLMs struggle with high-precision numerical analysis~\citep{hendrycksmath2021, hu2024case}, making it difficult to fine-tune an LLM to convey commonsense knowledge in a sufficiently precise physical form for manipulation tasks that demand high accuracy. \cref{fig:phy-knowledge-pipeline}-(a) illustrates the differences between knowledge in the semantic form learned by LLMs and the physical form understood by robots. 

In order to construct a bridge between the semantic-level knowledge inferred by LLMs and the physical world where real robots operate, we introduce analytic concepts. In our vision, a piece of commonsense knowledge encapsulates the essential commonality shared by a group of similar entities. From this perspective, an analytic concept is procedurally defined with mathematical symbols to represent such generalized commonality in a physical form which can be directly computed and simulated by machines. By aligning the semantic knowledge provided by LLMs with analytic concepts and mapping the analytic concepts to the physical world, we can identify important priors regarding object structure and functionality according to these concepts, and finally use these pieces of physically grounded knowledge to instruct robot control policies.

\cref{fig:phy-knowledge-pipeline}-(b) illustrates our improvement on current LLM-instructed paradigm using analytic concepts. Given an open-ended description of an object manipulation task in natural language and an RGB-D workspace image, a robot should perform appropriate physical interactions. In common practice (\cref{fig:phy-knowledge-pipeline}-(b)-Upper), a multi-modal LLM (MLLM) analyzes the task description and the image, reasons about where and how to interact with the target object, and outputs a semantic-level task plan to guide the control policy. 
In contrast (\cref{fig:phy-knowledge-pipeline}-(b)-Lower), we provide the MLLM not only with the task description and image but also with our proposed analytic concepts. With their designs (described in Sec.\cref{sec:analytic-concepts}), the MLLM can thoroughly understand and integrate these analytic concepts with other information through its commonsense reasoning capability. 
We then prompt the MLLM to provide responses by employing analytic concepts, focusing on three critical aspects: the target part for manipulation, its structural knowledge, and the manipulation knowledge required to complete the manipulation task. 
This approach aligns the semantic-level knowledge inferred by the MLLM with the analytic concepts, which are then grounded in the physical world. The control policy subsequently incorporates the knowledge within the grounded concepts with the RGB-D data of the target object to produce reliable interaction strategies guided by the embedded physical knowledge. 

By bridging the semantic knowledge and the physical world through analytic concepts, our approach greatly benefits from the capability of MLLMs in commonsense reasoning and instructing robot control policies for generalized, interpretable, and accurate articulated object manipulation. We have conducted exhaustive experiments in both simulation and real-world environments to demonstrate the superiority of our approach in articulated object manipulation. 

In summary, our contributions are as follows:  \textbf{1)} We introduce analytic concepts to build a bridge between semantic-level knowledge inferred by MLLMs and the physical world where robots operate. \textbf{2)} We propose a pipeline to ground semantic-level knowledge in the physical world through analytic concepts, enabling concrete and precise guidance for robots in completing various manipulation tasks. \textbf{3)} Benefiting from MLLMs' strong reasoning capability and analytic concepts as a physically grounded representation, our method demonstrates great superiority in articulated object manipulation tasks across extensive object categories in both simulation and real-world environments. 

\begin{figure*}[htbp]
    \centering
    \includegraphics[width=\linewidth]{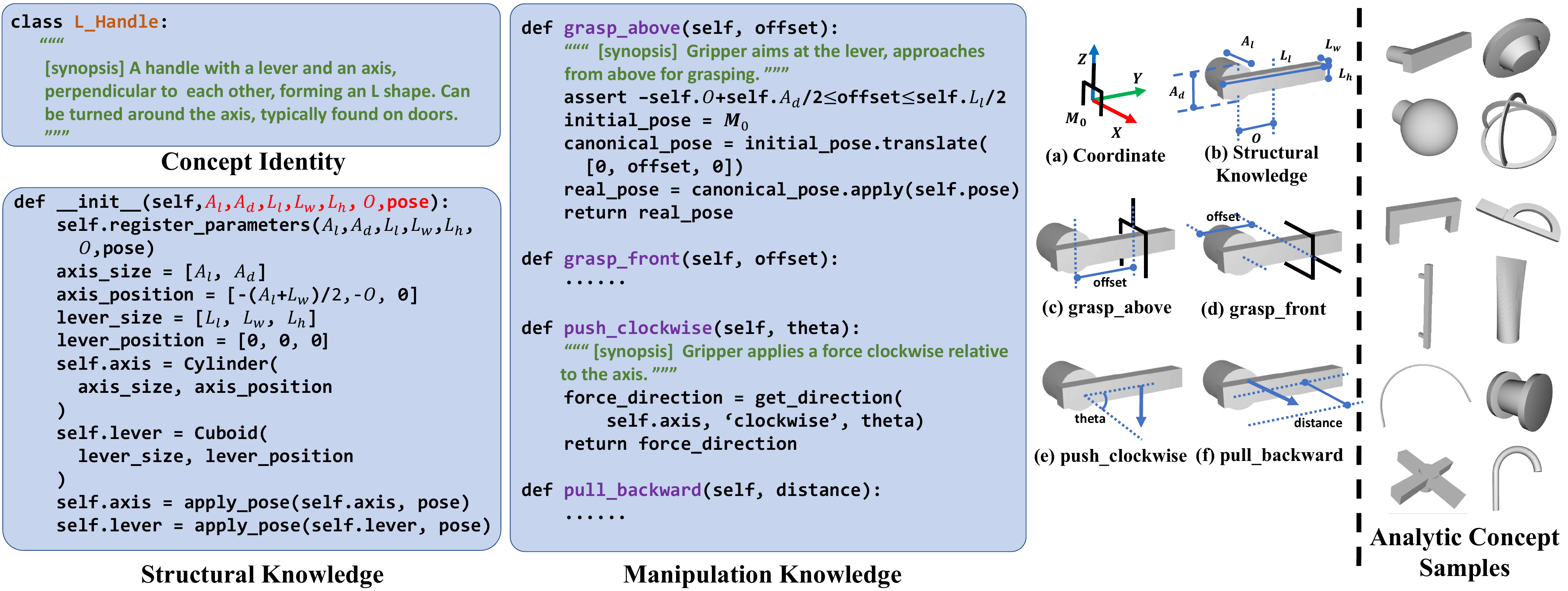}
    \caption{[Left] Example implementation of concept \texttt{\textbf{L\_Handle}}, including \textbf{Concept Identity}, \textbf{Analytic Structural Knowledge} defined in 3D space (a-b), and \textbf{Analytic Manipulation Knowledge} (c-f). [Right] Visualizations of representative analytic concepts. }
    \label{fig:analytic-concepts}
\end{figure*}

\section{Related Works}

\subsection{Articulated Object Manipulation}

Articulated object manipulation requires embodied agents to interact with objects using both visual perception and physical reasoning~\citep{mo2021where2act, geng2023gapartnet, li2024manipllm, huang2024manipvqa, qian2024affordancellm, huang2025a3vlm, pan2025omnimanip}. Researchers have approached this challenge from multiple directions. For example, \citet{xiang2020sapien} builds a physics-rich simulator paired with a diverse real-world object dataset. Where2Act~\citep{mo2021where2act} predicts per-pixel action likelihoods to generate targeted manipulation proposals. Where2Explore~\citep{ning2024where2explore} introduces a few-shot framework that measures affordance similarity across categories, enabling transfer of affordance knowledge to novel objects. GAPartNet~\citep{geng2023gapartnet} proposes generalized actionable parts and uses GAPart poses as structured representations to define heuristic interaction policies. ManipLLM~\citep{li2024manipllm} leverages LLMs to infer action policies directly from a single RGB image of the target object. A3VLM~\citep{huang2025a3vlm} further uses MLLMs to infer the actionable part’s bounding box, axis, and semantics, integrating them to derive manipulation policies.

\subsection{Multi-modal Large Language Models}

Recent large language models, such as LLaMA~\citep{touvron2023llama} and GPT-3~\citep{floridi2020gpt}, demonstrate strong capabilities in solving complex tasks, underscoring their potential to learn and apply commonsense knowledge. To better harness these capabilities for vision research, MLLMs~\citep{achiam2023gpt, hurst2024gpt, liu2024visual, lin2023sphinx} have been developed, significantly expanding visual understanding and processing. For example, GPT-4o~\citep{hurst2024gpt} integrates commonsense reasoning with multimodal inputs to interpret and respond to visual tasks, achieving competitive performance in areas such as image captioning and visual question answering. 

However, applying MLLMs to object manipulation tasks~\citep{li2024manipllm, qian2024affordancellm, huang2024manipvqa, pan2025omnimanip,qi2025sofar,huang2025rekep}, which require precise physical interactions, remains challenging due to the need to align semantic, visual, and action representations with their grounding in the physical world~\citep{ma2024survey}. In this paper, we introduce analytic concepts to more effectively leverage the strengths of MLLMs in physical environments and enable more generalizable robotic manipulation skills. 

\section{Analytic Concepts}
\label{sec:analytic-concepts}
In this section, we introduce how analytic concepts are developed to represent commonsense knowledge in a physical form. As shown in \cref{fig:analytic-concepts}-Left, each analytic concept comprises three components, concept identity, analytic structural knowledge, and analytic manipulation knowledge. Since manipulation knowledge is linked with a specific spatial structure, we focus primarily on structural knowledge when designing a concept and integrate the related manipulation knowledge within it.

\subsection{Concept Identity}

Each concept is assigned a unique identity as a symbol representing specific knowledge, ensuring one-to-one correspondence. In addition, a concise and precise synopsis is provided alongside the identity, enabling the concept to be interpreted consistently by both humans and MLLMs. 

\subsection{Analytic Structural Knowledge}

Analytic structural knowledge uses basic geometries (\textit{e.g.}, \textit{Cylinder}, \textit{Cuboid}, \textit{Sphere}, \textit{etc.}) as fundamental structural elements and represents a spatial structure by arranging and combining them through mathematical procedures to capture the commonalities of a specific structure, including the layout and structural relationships. These procedures incorporate variable parameters to represent variations among different instances, enabling the spatial structure of actionable part to be described using an analytic concepts for grounding its structural knowledge. 

\subsection{Analytic Manipulation Knowledge}

Interaction with objects relies on understanding its physical properties, including affordances and physical effects of applied forces. To provide such properties, analytic manipulation knowledge specifies grasp poses aligned with the concept’s structural knowledge and force directions that produce effective motion, which are formulated by applying mathematical procedures to the variable parameters of the corresponding structural knowledge. As illustrated in \cref{fig:analytic-concepts}-Left-(c–f), each concept may contain multiple manipulation knowledge, with each piece represented by a function that can be parameterized into a specific grasp pose or force direction. This design covers all atomic actions applicable to the concepts and introduces diversity through parameter variations. Additionally, a concise synopsis is provided for each piece to support reasoning by MLLMs as the bridge from semantic-level knowledge to physical level.

\subsection{Scalability and Generalizability} 
\label{subsec:analytic-concept-discussion}

Analytic concepts capture commonsense knowledge for object interaction for bridging the semantic-level knowledge to physical levels, requiring them to be easy to understand, clearly defined, and broadly applicable. We invite volunteers with high-school-level math skills to create analytic concepts, along with their associated knowledge and synopses, and experts check the results for correctness. For each concept, different volunteers produce consistent templates to represent the structure and knowledge, indicating that the definition of analytic concepts shows consistency in cognition level. In addition, each concept can be completed in 2 hours on average, showing the \textbf{scalability} to create analytic concepts. So far, we have created 153 analytic concepts, and we find that only a small subset is sufficient to support a wide range of object manipulation tasks, indicating the strong \textbf{generalizability} of analytic concepts. \cref{fig:analytic-concepts}-Right shows visualization of selected implementation of analytic concepts. 

\begin{figure*}[t]
    \centering
    \includegraphics[width=\linewidth]{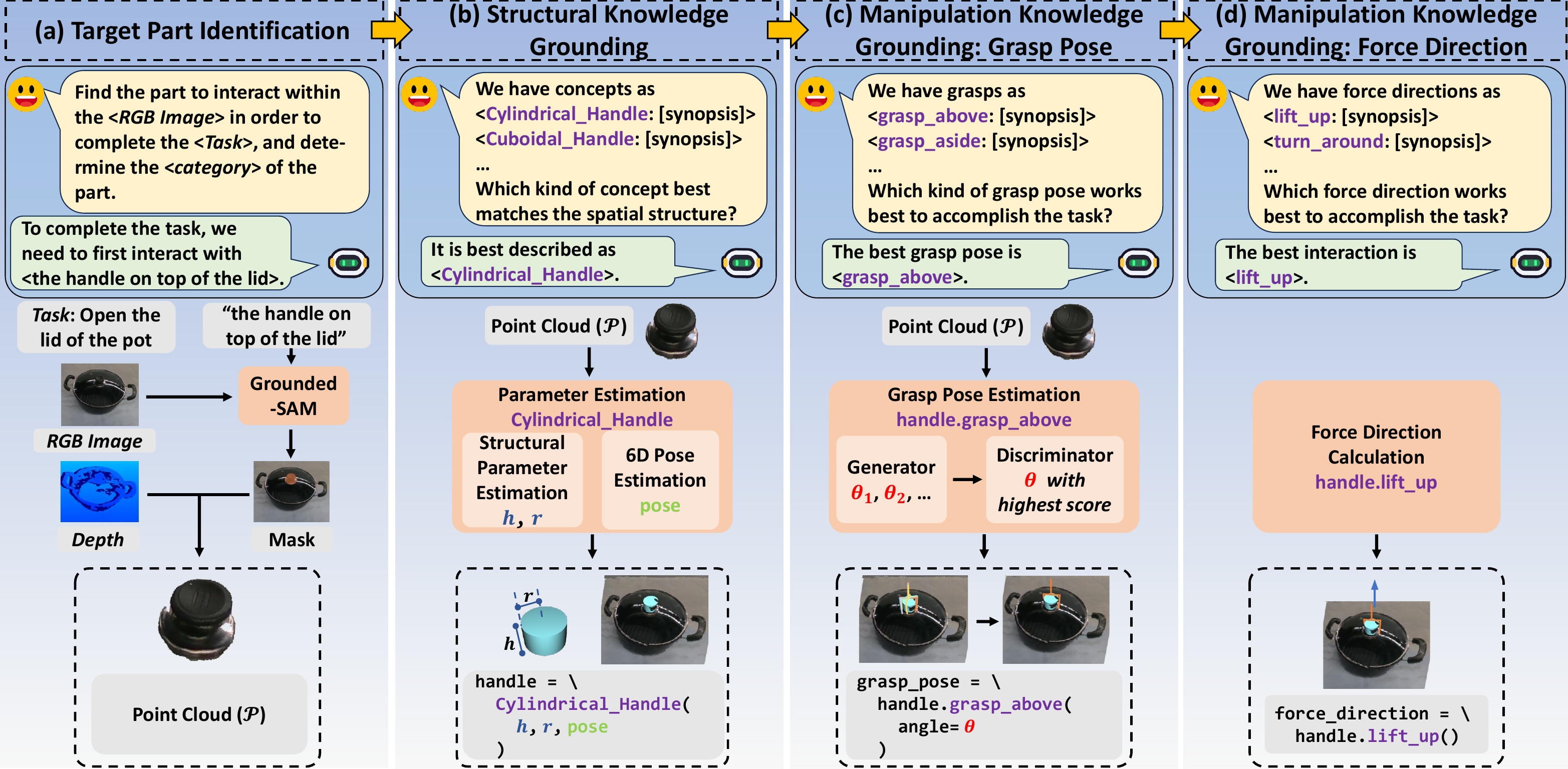}
    \caption{Demonstration of our approach consists of three major steps: (a) Target Part Identification, (b) Structural Knowledge Grounding, and (c–d) Manipulation Knowledge Grounding. Each step begins with a question-answering process, in which the MLLM infers semantic-level knowledge in the downstream processes. }
    \label{fig:full-pipeline}
\end{figure*}

\section{Method}

In this section, we describe our method to ground semantic-level knowledge reasoned by MLLMs into the physical world with analytic concepts, and show the benefits physically grounded knowledge brings to object manipulation. 

\paragraph{Problem Formulation.} Given an RGB-D image of an object, along with a task description in natural language, a robot is required to perform appropriate physical interactions using a parallel gripper to accomplish the task. As shown in \cref{fig:full-pipeline}, we propose a pipeline to leverage analytic concepts as the bridge between semantic-level knowledge to physical-level knowledge with three major steps: 1) target part identification, 2) structural knowledge grounding, and 3) manipulation knowledge grounding. 

\subsection{Target Part Identification} 
\label{subsec:target-part}

We use an MLLM to identify the target part for manipulation from the task definition and the RGB image. Specifically, we employ GPT-4o~\citep{hurst2024gpt} and prompt it with: ``\textit{find the \textbf{part} to interact with in the $<$RGB Image$>$ in order to complete the task $<$Task$>$, and determine the $<$category$>$ of the \textbf{part}}." to acquire both a semantic-level description and the category of the target part. The semantic-level description along with the RGB image is then passed to Grounded-SAM~\citep{ren2024grounded} to detect a pixel-level segmentation of the part. The segmentation mask is applied to the depth image to crop the point cloud $\mathcal{P}$ of the target part, while the category information is retained for identifying the corresponding analytic concept. 

\subsection{Structural Knowledge Grounding}
\label{subsec:ground-structual-knowledge}

After obtaining the point cloud $\mathcal{P}$ of the target part being manipulated, we further leverage the MLLM to reason about the structural knowledge of this part and ground it on the corresponding analytic concept in physical level with parameter estimators. 

\vspace{-6pt}

\paragraph{Concept Identification.} 
Leveraging the commonsense reasoning capability of the MLLM, we identify the analytic concept that best represents the spatial structure of the target part. The analytic concepts are grouped (\textit{e.g.}, handle, lid) in advance based on the spatial structures they describe and their synopses. To select the best-matching concept, we provide the MLLM with the identities and synopses of concepts within the same group as the target part, and prompt it to determine ``\textit{which \textbf{concept} best matches the spatial structure of the target part}." This procedure aligns the MLLM’s semantic-level understanding with the analytic concepts. 

\vspace{-6pt}

\paragraph{Parameter Estimation.} 
After the concept is identified, its parameters are estimated to ground the concept onto the target part. These parameters fall into two categories: (i) structural parameters that define the spatial structure specified in the concept’s analytic formulation, and (ii) 6-DoF pose parameters that describe its global translation and rotation in world coordinates. The estimation procedure is therefore carried out in two corresponding stages. 

For the structural parameters, a Point-Transformer~\citep{zhao2021point} encoder is applied to encode the target point cloud $\mathcal{P}$ into latent space, followed by separate MLP heads that regress the parameters for each analytic concept type. 

For the 6-DoF pose parameters, the point cloud $\mathcal{P}$ is encoded using an encoder~\citep{zhao2021point} followed by an MLP decoder, which outputs a point cloud $\mathcal{P}^*$ representing $\mathcal{P}$ in the canonical space of the analytic concept. The Umeyama algorithm~\citep{umeyama1991least} combined with RANSAC~\citep{fischler1981random} for outlier removal is then applied to estimate the rigid transformation $\mathbf{T}\in SE(3)$ from $\mathcal{P}^*$ to $\mathcal{P}$. The resulting translation and rotation constitute the 6-DoF pose of the analytic concept. 

\subsection{Manipulation Knowledge Grounding}

\label{subsec:ground-manip-knowledge}

The physical-level grounding of analytic structural knowledge enables the grounding of manipulation knowledge, including grasp poses and force directions that guide the robot in executing the manipulation task at physical level. The process begins by prompting the MLLM with the available grasp-pose and force-direction knowledge, along with their synopses, and asking it to determine ``\textit{which \textbf{grasp pose} / \textbf{force direction} works best to accomplish the task}."

\paragraph{Grasp Pose.} 

Each piece of analytic manipulation knowledge for grasp poses (\textit{e.g.}, grasp above, grasp in front, \textit{etc.}) defines a class of poses possessing a shared pattern, with varying parameters specifying a particular grasp pose $\mathbf{G}$. 

To estimate these parameters, a generative framework based on conditional GANs~\citep{mirza2014conditional} is adopted. The generator $G$ employs MLPs to produce multiple candidate parameters from Gaussian noise $z$, conditioned on features of the target point cloud $\mathcal{P}$ extracted with an encoder~\citep{zhao2021point}. A discriminator $D$ then scores each candidate to select the most suitable one. The discriminator encodes $\mathcal{P}$ and the parameters, concatenates the encoded features, and decodes them via MLPs to produce a score in $(0,1)$. Using the selected parameters, a physically grounded grasp pose $\mathbf{G}$ can be mathematically defined according to the analytic manipulation knowledge. 

\paragraph{Force Direction.} 
The force direction $\mathcal{F}$ (\textit{e.g.} lift up, turn clockwise, \textit{etc.}) specifies the direction in which the robot should apply force to the target part. It is mathematically defined by the parameters of the structural knowledge and the grasp pose. The force direction for the task is selected by the MLLM, and is then procedurally computed. 

\subsection{Object Interaction}

Guided by the physically grounded knowledge embedded in analytic concepts, a robot can interact with the object heuristically. The robot is guided to move its end-effector to grasp the object according to $\mathbf{G}$, and moves along the force direction $\mathcal{F}$ to manipulate the object.

\subsection{Implementation} 

\paragraph{Grounded-SAM.} 
Grounded-SAM~\citep{ren2024grounded} consists of two major components, Grounding-Dino~\citep{liu2024grounding} and SAM~\citep{kirillov2023segment}. We keep SAM~\citep{kirillov2023segment} frozen and fine-tune Grounding-Dino~\citep{liu2024grounding} with RGB images of size $336\times 336$ with ground-truth bounding boxes of the actionable parts, along with natural language prompt that describes the actionable parts provided by GPT-4o~\citep{hurst2024gpt}.

\vspace{-6pt}

\paragraph{Structural Parameter Estimation.}
The encoder is a Point-Transformer~\citep{zhao2021point} that extracts 128 groups of points with size 32 from the input with 2048 points and has 12 6-headed attention layers. The subsequent MLP with three layers outputs the structural parameters. The network is trained with L2 loss between the estimated and ground-truth structural parameters.

\vspace{-6pt}

\paragraph{6-DoF Pose Estimation.}
The network shares the same architecture as that used in structural parameter estimation. The network is trained with chamfer distance loss between the estimated point cloud and the point cloud of the input in canonical space which is obtained by transforming the input point cloud according to its ground-truth 6-DoF pose.

\vspace{-6pt}

\paragraph{Grasp Pose Estimation.} 
The Point-Transformer~\citep{zhao2021point} and MLPs in both generator and discriminator shares the same architecture as that used in structural parameter estimation. As we already have positive and negative grasp pose parameter samples, we first train the discriminator with $\mathcal{L}_D = -\mathbb{E}_{x \sim p_{\text{data+}}} \left[\log D(x|y)\right] - \mathbb{E}_{x \sim p_{\text{data-}}} \left[\log \left(1 - D(x|y)\right)\right]$, where $D(x|y)$ is the output score of the discriminator representing the likelihood of sample $x$ in the data distribution given condition $y$, and $p_{\text{data+}}$ and $p_{\text{data-}}$ denotes the positive and negative data distribution. Then we train the generator with $\mathcal{L}_G = -\mathbb{E}_{z \sim p_z} \left[\log D(G(z|y))\right]$ where $G(z|y)$ refers to the generated sample according to noise $z$ given condition $y$, and $p_z$ is a Gaussian distribution.

\section{Experiments}

We thoroughly evaluate our approach on articulated object manipulation tasks against five representative approaches~\citep{mo2021where2act,geng2023gapartnet,ning2024where2explore,li2024manipllm,huang2025a3vlm}, including state-of-the-art A3VLM~\citep{huang2025a3vlm} to showcase the effectiveness of our approach. In the following sections, we present the simulation setting, experiment results and comprehensive analysis. We also carry out the object manipulation experiments with physical robots in real-world environments to provide a more comprehensive and stronger evaluation. 

\begin{table*}[htbp]
    \centering
    \caption{Evaluation on manipulation tasks across different object categories. All values represent the average success rate as a percentage. The average success rates are calculated based on the objects. The full name of the object categories are listed as follows: Box, Door, Faucet, Fridge, Kettle, Microwave, StorageFurniture, Switch, TrashCan, Window, Bucket, KitchenPot, Safe, Table, WashingMachine. }
    \label{tab:main_res}
    \resizebox{\textwidth}{!}{
    \begin{tabular}{cccccccccccccccccccc}
    \toprule
        \multirow{2}*{Methods} & & \multicolumn{11}{c}{\cellcolor{green!20}\textbf{Training Categories}} & & \multicolumn{6}{c}{\cellcolor{blue!20}\textbf{Testing Categories}} \\
        \cmidrule{3-13} \cmidrule{15-20}
        & & Box & Dor & Fct & Fdr & Ket & Mcw & Stf & Swt & Tcn & Win & \textbf{AVG} & & Bkt & Pot & Saf & Tab & Wsm & \textbf{AVG} \\ 
        \cmidrule{1-1} \cmidrule{3-13} \cmidrule{15-20}
        Where2Act~\citep{mo2021where2act} & & 6.8 & 31.5 & 17.1 & 31.2 & 9.9 & 31.3 & 37.1 & 19.3 & 17.1 & 12.2 & 26.1 & & 8.0 & 6.2 & 20.4 & 17.8 & 9.1 & 14.4 \\ 
        Where2Explore~\citep{ning2024where2explore} & & 9.0 & 38.3 & 16.8 & 35.4 & 11.5 & 33.8 & 34.6 & 23.6 & 19.5 & 15.8 & 26.9 & & 14.0 & 10.8 & 31.8 & 22.7 & 15.4 & 20.5 \\
        GAPartNet~\citep{geng2023gapartnet} & & 15.5 & 45.4 & 17.1 & 40.5 & 11.1 & 35.8 & 40.5 & 18.2 & 19.5 & 13.8 & 29.7 & & 21.3 & 14.7 & 30.2 & 37.3 & 12.7 & 28.7 \\ 
        ManipLLM~\citep{li2024manipllm} & & 11.0 & 50.2 & 16.5 & 36.4 & 10.6 & 41.5 & 47.0 & 20.4 & 15.0 & 15.2 & 32.0 & & 19.9 & 12.4 & 37.0 & 39.1 & 19.1 & 30.6 \\ 
        A3VLM~\citep{huang2025a3vlm} & & 12.2 & 56.2 & 24.1 & 40.8 & 20.1 & 49.7 & 53.3 & 25.4 & 18.2 & 15.8 & 37.4 & & 25.2 & 11.0 & 40.0 & 42.5 & 22.3 & 32.1 \\ 
        Ours & & \textbf{15.9} & \textbf{58.4} & \textbf{28.0} & \textbf{46.7} & \textbf{24.3} & \textbf{52.2} & \textbf{58.6} & \textbf{32.7} & \textbf{22.9} & \textbf{22.3} & \textbf{42.5} & & \textbf{28.9} & \textbf{15.5} & \textbf{50.6} & \textbf{51.6} & \textbf{22.7} & \textbf{40.8} \\ 
        \bottomrule
    \end{tabular}
    }
\end{table*}

\begin{table}[t]
    \centering
    \caption{Results of real-world experiments. All values represent success rate. }
    \label{tab:real-exp}
    \resizebox{\linewidth}{!}{
    \begin{tabular}{ccccccccc}
    \toprule
         \raisebox{1mm}{Method} & \includegraphics[width=0.08\linewidth]{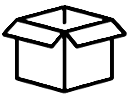} & \includegraphics[width=0.08\linewidth]{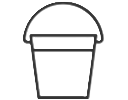} & \includegraphics[width=0.08\linewidth]{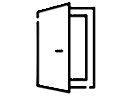} & \includegraphics[width=0.08\linewidth]{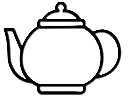} & \includegraphics[width=0.08\linewidth]{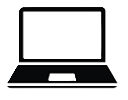} & 
         \includegraphics[width=0.08\linewidth]{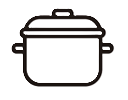} & 
         \includegraphics[width=0.08\linewidth]{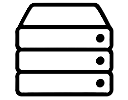} & 
         \includegraphics[width=0.08\linewidth]{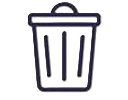}\\
         \midrule
         A3VLM~\citep{huang2025a3vlm} & 0.70 & 0.40 & 0.50 & 0.60 & 0.70 & 0.60 & 0.70 & 0.60 \\
         Ours & 0.90 & 0.60 & 0.70 & 0.80 & 0.90 & 0.80 & 0.80 & 0.70 \\
         \bottomrule
    \end{tabular}
    }
\end{table}

\subsection{Simulation Evaluation}

\subsubsection{Simulation Settings}
\label{subsec:simulation-setting}

We design our evaluation in order to reflect the capability of an algorithm to accurately guide a robot to interact with articulated objects with a single gripper. Hence, we generally follow the experiment setting of Where2Act~\citep{mo2021where2act} which is widely used in the research community. 

\paragraph{Data Settings.} 
Since our manipulation knowledge is developed for single-gripper manipulation, 
We have collected a total of 972 objects across 15 categories from PartNet-mobility~\citep{xiang2020sapien} that are suitable for single-gripper manipulation following \citep{mo2021where2act}. To evaluate the generalization of our approach to novel categories, we divide the categories into a training group with 10 categories and a testing group with 5 categories. 

\paragraph{Experiment Settings.} 
We adopt the SAPIEN~\citep{xiang2020sapien} simulator as the simulation environment for our evaluation. In each manipulation simulation, the target object is initially placed at the center of the scene within the simulator. The object's joint pose is randomized, with a 50\% chance of being at the closed state and a 50\% chance at open state with random motion. An RGB-D camera with known intrinsic parameters stares at the center of the scene and is positioned at the upper hemisphere with a random azimuth $[0^\circ,360^\circ)$ and a random elevation $[30^\circ,60^\circ]$. We use a gripper as the end-effector in our main experiments following~\citep{mo2021where2act,ning2024where2explore}.  

\paragraph{Evaluation Settings.} 
We use the success rate as our evaluation metric to assess the capability of articulated object manipulation among different approaches. A manipulation is considered successful if the target joint's movement exceeds 0.01 unit-length or 0.5 relative to its maximum motion range. The success rate is defined as the ratio of successful manipulations to the total number of test trials. We adopt an interaction budget of 5 for each action proposal.

\subsubsection{Simulation Results and Analysis}
\label{sec-simulation-evaluation}

We compare our approach with four baselines representing three types of frameworks for articulated object manipulation. Vanilla frameworks, such as Where2Act~\citep{mo2021where2act} and Where2Explore~\citep{ning2024where2explore}, learn directly from 2D images or point clouds and output pixel-level affordance maps. The framework leveraging object structures such as GAPartNet~\citep{geng2023gapartnet}, takes point clouds as input but proposes the idea of GAPart, treating the 6-DoF poses as a kind of structured representation, and adopts the interaction policies predefined on each GAPart. The third type of framework, including ManipLLM~\citep{li2024manipllm} and A3VLM~\citep{huang2025a3vlm}, leverages the robust reasoning capabilities of MLLMs to infer affordance knowledge from RGB-D image and natural language inputs featuring a target object. In comparison with these frameworks, our approach integrates both the MLLM's reasoning capability and analytic concepts as a physically grounded structured representation for object manipulation tasks. 

\vspace{-9pt}

\paragraph{Main Results.} 
The performance of different baselines in \cref{tab:main_res} highlights the critical role of commonsense reasoning and physically grounded representations in object manipulation tasks. Compared to vanilla frameworks Where2Act~\citep{mo2021where2act} and Where2Explore~\citep{ning2024where2explore}, both ManipLLM~\citep{li2024manipllm} and A3VLM~\citep{huang2025a3vlm} enhance one of these two aspects and achieve certain improvements. After fully integrating these two aspects through our approach by grounding commonsense knowledge inferred by MLLMs on analytic concepts, the success rate of manipulation has been boosted, about 15.2\% for train categories and 27.1\% for test categories compared with A3VLM~\citep{huang2025a3vlm}, demonstrating the superiority of our proposed analytic concepts and framework. Notably, for categories consisting of objects with complex structures and multiple joints, such as Table, the success rate of our approach has been boosted 21.4\% compared with A3VLM~\citep{huang2025a3vlm}. In addition, the close results on train and test categories also suggest that our approach can effectively handle unseen objects. These results further demonstrate the strength of our approach, which leverages the commonsense knowledge of MLLM to infer the interaction target and manipulation knowledge for complex objects, and grounds the knowledge to physical world using analytic concepts for accurate manipulation. We attribute this property to the effect of the following two factors. First, our defined analytic concepts can cover ubiquitous common structure and generalize effectively to describe unseen objects. Second, utilizing its strong commonsense reasoning capability, an MLLM can still find the best-match concepts to the target parts of unseen objects according to the concepts' synopses.

\begin{figure}[t]
    \centering
    \includegraphics[width=\linewidth]{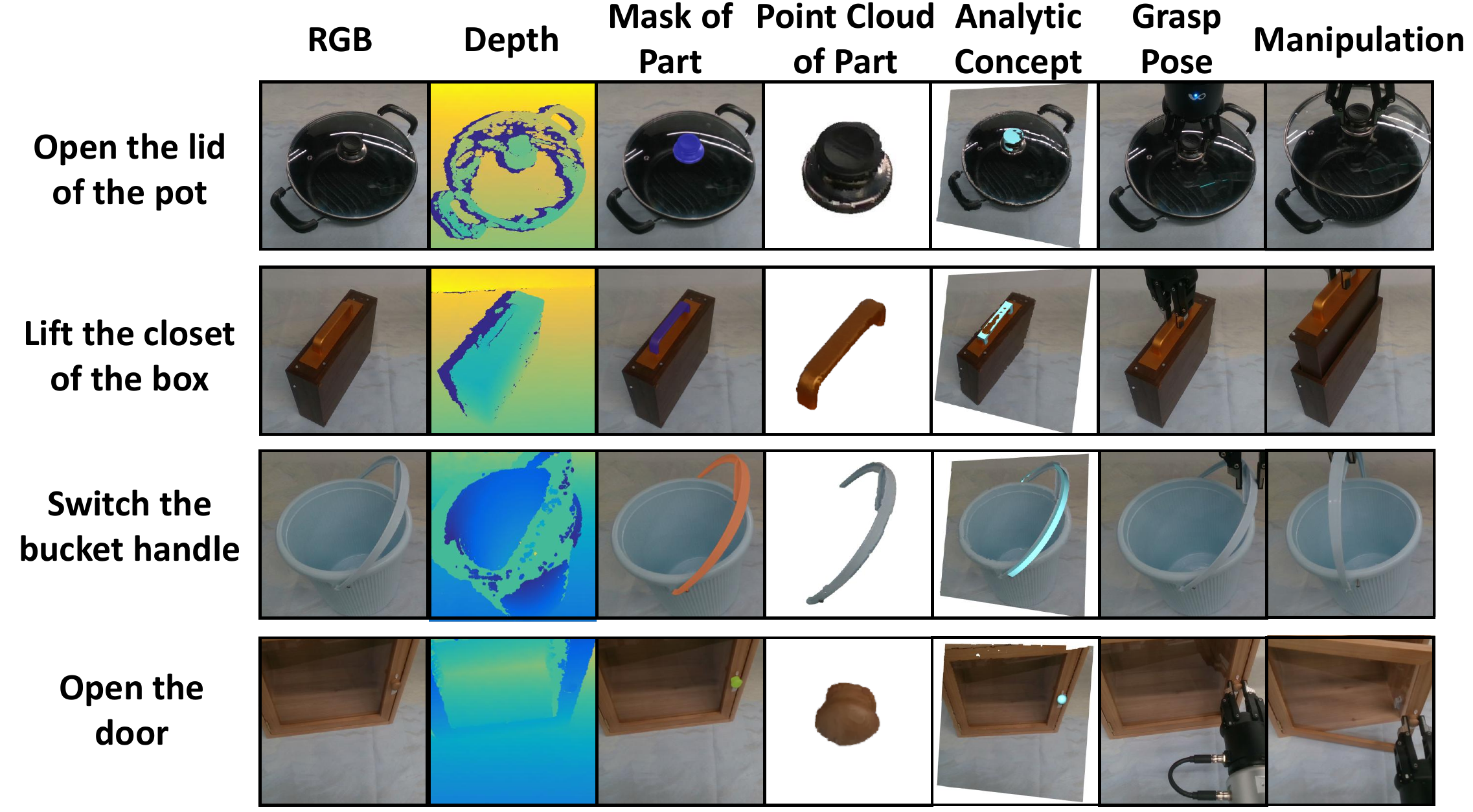}
    \caption{Visualizations of the output in each step in real-world experiments. The leftmost column lists the task description. In the \textbf{Analytic Concept} column, light blue meshes represent analytic concepts grounded on the target parts. The \textbf{Grasp Pose} and \textbf{Manipulation} columns illustrate how the gripper grasps the target parts and completes the manipulation tasks.} 
    \label{fig:qualitative}
\end{figure}

\begin{table*}[t]
    \centering
    \caption{Comparison between ManipLLM, A3VLM and our approach on manipulation tasks using and \textbf{suction} as the end effector. All values represent the average success rate in percentage.}
    \label{tab:cmp_manipllm}
    \resizebox{\textwidth}{!}{
    \begin{tabular}{cccccccccccccccccccc}
    \toprule
        \multirow{2}*{\makecell{Suction-based\\Methods}} & & \multicolumn{11}{c}{\cellcolor{green!20}\textbf{Training Categories}} & & \multicolumn{6}{c}{\cellcolor{blue!20}\textbf{Testing Categories}} \\
        \cmidrule{3-13} \cmidrule{15-20}
        & & Box & Dor & Fct & Fdr & Ket & Mcw & Stf & Swt & Tcn & Win & \textbf{AVG} & & Bkt & Pot & Saf & Tab & Wsm & \textbf{AVG} \\ 
        \cmidrule{1-1} \cmidrule{3-13} \cmidrule{15-20}
        ManipLLM~\citep{li2024manipllm} & & 62.9 & 61.7 & 44.7 & 74.2 & 46.9 & 62.5 & 68.5 & 41.1 & 50.6 & 40.2 & 58.0 & & 44.3 & 52.4 & 58.9 & 65.5 & 47.9 & 57.8 \\ 
        A3VLM~\citep{huang2025a3vlm} & & 75.8 & 72.9 & 63.8 & 79.4 & 70.0 & 76.7 & 86.1 & 44.3 & 70.2 & 47.1 & 72.4 & & 58.2 & 65.3 & 79.3 & 68.5 & 50.6 & 66.4 \\ 
        Ours & & \textbf{79.3} & \textbf{80.2} & \textbf{65.5} & \textbf{82.5} & \textbf{73.1} & \textbf{84.7} & \textbf{88.1} & \textbf{50.4} & \textbf{72.6} & \textbf{49.8} & \textbf{75.5} & & \textbf{61.0} & \textbf{70.3} & \textbf{81.4} & \textbf{80.8} & \textbf{52.8} & \textbf{73.8} \\ 
        \bottomrule
    \end{tabular}
    }
\end{table*}

\begin{table*}[t]
    \centering
    \caption{Evaluation results with grasp pose parameters randomly sampled in parameter space and estimated by our proposal. All values represent the average success rate as a percentage. }
    \label{tab:afford_retr_res}
    \resizebox{\textwidth}{!}{
    \begin{tabular}{cccccccccccccccccccc}
        \toprule
        \multirow{2}*{\makecell{Manipulation\\Knowledge}} & & \multicolumn{11}{c}{\cellcolor{green!20}\textbf{Training Categories}} & & \multicolumn{6}{c}{\cellcolor{blue!20}\textbf{Testing Categories}} \\
        \cmidrule{3-13} \cmidrule{15-20}
        & & Box & Dor & Fct & Fdr & Ket & Mcw & Stf & Swt & Tcn & Win & \textbf{AVG} & & Bkt & Pot & Saf & Tab & Wsm & \textbf{AVG} \\ 
        \cmidrule{1-1} \cmidrule{3-13} \cmidrule{15-20}
        Sampled & & 14.8 & 58.0 & 26.5 & 42.9 & 23.3 & 50.3 & 55.9 & 30.0 & 20.6 & 20.2 & 40.2 & & 28.0 & 14.0 & 47.1 & 48.7 & 22.7 & 38.6 \\ 
        Estimated & & \textbf{15.9} & \textbf{58.4} & \textbf{28.0} & \textbf{46.7} & \textbf{24.3} & \textbf{52.2} & \textbf{58.6} & \textbf{32.7} & \textbf{22.9} & \textbf{22.3} & \textbf{42.5} & & \textbf{28.9} & \textbf{15.5} & \textbf{50.6} & \textbf{51.6} & \textbf{22.7} & \textbf{40.8} \\ 
        \bottomrule
    \end{tabular}
    }
\end{table*}

\subsection{Real-world Evaluation}
\label{subsec:real-world}

\paragraph{Settings.} 
We conduct experiments that involve interacting with 8 real-world household objects from different categories. The manipulation tasks involve \textit{open the box}, \textit{switch the handle of the bucket}, \textit{open the door}, \textit{lift the lid of the kettle}, \textit{open/close the laptop}, \textit{lift the lid of pot}, \textit{open the closet of the storage} and \textit{move the lid of the trashcan}. We employ a robot arm with a parallel gripper as its end-effector and utilize a RealSense D415 camera to capture the RGB-D image. We adopt success rate as metric, where we define a successful interaction as one in which the gripper securely grasps the target object and moves correctly to complete the task. Each task is tested for 10 times. 

\vspace{-6pt}

\paragraph{Quantitative Analysis.} 
\cref{tab:real-exp} shows the performance of our approach in real-world experiments, demonstrating high success rates. The comparison with A3VLM~\citep{huang2025a3vlm} highlights the importance of a physically grounded knowledge representation in accurate and reliable robot manipulation tasks.

\vspace{-6pt}

\paragraph{Qualitative Analysis.}
\cref{fig:qualitative} visualizes the outputs of each step in four real-world manipulation tasks, demonstrating our approach's ability to accurately locate the target part, physically ground both structural and manipulation knowledge and successfully complete the manipulation tasks. We attribute this success to the effective integration of both the strong reasoning capability of MLLMs and the physically grounded knowledge represented by analytic concepts. 

\subsection{Ablation Studies}
\label{subsubec-ablation-studies}

\paragraph{Analytic Concept on Suction.} 
ManipLLM~\citep{li2024manipllm} and A3VLM~\citep{huang2025a3vlm} adopt LLaMa-Adapter~\citep{zhang2023llama} as the MLLM backbone to infer how an end-effector interacts with the target object, where suctions are originally implemented. Although our main experiments use a parallel gripper (results shown in \cref{tab:main_res}), we additionally evaluate the performance of our method using suction as the end-effector under our data setting. We modify the analytic manipulation knowledge according to the properties of suctions. As shown in \cref{tab:cmp_manipllm}, the results using suction also demonstrate the effectiveness of our approach. In addition, after changing the end-effector from suction to parallel gripper, ManipLLM~\citep{li2024manipllm} and A3VLM~\citep{huang2025a3vlm} both suffer a more severe drop in success rate compared with our approach. This highlights the challenge of relying solely on MLLMs to directly perform accurate numerical reasoning to produce precise affordances and interaction policies for manipulation tasks involving gripper-based grasping and interactions. In comparison, our approach is able to ground the semantic-level knowledge inferred by MLLMs through analytic concepts. In this manner, a robot can precisely calculate the affordances and interaction strategies with these physically grounded concepts, and achieves higher success rates, especially for more challenging settings, \textit{i.e.} gripper as end-effector for grasping. 

\paragraph{Effectiveness of Grasp Pose Knowledge.} 

We include a control group in which the grasp pose parameters are randomly \textit{sampled} within the grasp pose parameter space, rather than being \textit{estimated} by the grasp pose generator and discriminator networks. The evaluation results are shown in \cref{tab:afford_retr_res}. The comparison between \textit{sampled} parameters and \textit{estimated} parameters demonstrates that our grasp pose knowledge in analytic concepts can precisely reveal the possible grasps for the robot gripper to interact with the target object. After employing our proposed module to incorporate the visual features (point cloud $\mathcal{P}$ of the target part) for parameter estimation instead of relying on random sampling, the success rate of manipulations can be further improved, which demonstrates the effectiveness of our proposed grasp pose knowledge.

\paragraph{System Limitation Analysis.} In this section, we provide a system error breakdown analysis by sequentially replacing the output of each module with ground truth to identify common failure cases and system limitations. Because ground truth is required for replacement, the analysis is only conducted on training categories. The results are reported in \cref{tab:system-breakdown-brief}. The experiments demonstrate that the primary bottlenecks in the system are structural parameter estimation and 6-DoF pose estimation, where a 20.8\% and 14.3\% gap on average is observed by replacing these two modules respectively. Errors in estimating an object’s spatial structure and 6-DoF pose can lead to grasp poses with collisions or misaligned grasps, and preventing the gripper from accurately detecting the target part to complete the manipulation task. 

\begin{table}[t]
    \centering
    \caption{Results of manipulation on system limitations. All values represent the average success rate as a percentage. }
    \label{tab:system-breakdown-brief}
    \resizebox{0.7\linewidth}{!}{
    \begin{tabular}{cc}
        \toprule
         Module & \textbf{AVG} ssr \\
         \midrule
         None & 42.5 \\ 
         Actionable Part Segmentation & 49.8 \\ 
         Concept Identification & 51.2 \\ 
         Structural Param Estimation& 72.0 \\ 
         6-DoF Pose Estimation & 86.3 \\ 
         Grasp Pose Estimation & 93.6 \\ 
         Force Direction & 98.6 \\ 
         \bottomrule
    \end{tabular}
    }
\end{table}

\section{Conclusion}

In this paper, we focus on articulated object manipulation and introduce analytic concepts to represent commonsense knowledge involved in this task in a physical form. Each concept encompasses an identity, as well as analytic structural and manipulation knowledge procedurally defined with mathematical symbols. These concepts construct a bridge between the semantic-level knowledge and the physical world, helping to overcome the limitations of MLLMs in precise numerical reasoning and the shortcomings of natural language in describing physical concepts. Taking advantage of analytic concepts, we further propose an object manipulation pipeline to successfully translate semantic-level commonsense knowledge inferred by MLLMs into concrete physical-level knowledge, instructing robots to complete various manipulation tasks accurately. Our approach demonstrates its superiority across a wide range of object categories in both simulation and real-world scenarios.

{
    \small
    \bibliographystyle{ieeenat_fullname}
    \bibliography{arxiv}
}

\end{document}